%% file: main.tex
\documentclass[sigconf,natbib]{acmart}

\AtBeginDocument{%
  }

\copyrightyear{2023}
\acmYear{2023}
\setcopyright{acmlicensed}\acmConference[SIGIR '23]{Proceedings of the 46th
International ACM SIGIR Conference on Research and Development in
Information Retrieval}{July 23--27, 2023}{Taipei, Taiwan}
\acmBooktitle{Proceedings of the 46th International ACM SIGIR Conference on
Research and Development in Information Retrieval (SIGIR '23), July 23--27,
2023, Taipei, Taiwan}
\acmPrice{15.00}
\acmDOI{10.1145/3539618.3592064}
\acmISBN{978-1-4503-9408-6/23/07}

\usepackage{paralist}
\usepackage{multirow}
\usepackage{enumerate}

\usepackage{balance}

\def\AA{\mathbf{A}}

\begin{document}

\title{Sinkhorn Transformations for Single-Query Postprocessing in Text-Video Retrieval}

\author{Konstantin Yakovlev}
\email{yakovlev.konstantin1@huawei-partners.com}
\orcid{0009-0005-9397-6081}

\affiliation{%
  \institution{Huawei Noah's Ark Lab}
  \city{Moscow}
  \country{Russia}
}

\author{Gregory Polyakov}
\email{polyakovgregory@huawei-partners.com}
\orcid{0009-0003-7536-9670}

\affiliation{%
  \institution{Huawei Noah's Ark Lab}
  \city{Moscow}
  \country{Russia}
}

\author{Ilseyar Alimova}
\email{ilseyar.alimova@huawei.com}
\orcid{0000-0003-4528-6631}

\affiliation{%
  \institution{Huawei Noah's Ark Lab}
  \city{Moscow}
  \country{Russia}
}

\author{Alexander Podolskiy}
\email{podolskiy.alexander@huawei.com}
\orcid{0000-0002-2892-7356}

\affiliation{%
  \institution{Huawei Noah's Ark Lab}
  \city{Moscow}
  \country{Russia}
}

\author{Andrey Bout}
\email{bout.andrey@huawei.com}
\orcid{0009-0007-8786-2269}

\affiliation{%
  \institution{Huawei Noah's Ark Lab}
  \city{Moscow}
  \country{Russia}
}

\author{Sergey Nikolenko}
\email{sergey@logic.pdmi.ras.ru}
\orcid{0000-0001-7787-2251}

\affiliation{%
  \institution{Ivannikov Institute for System Programming of the RAS}
  \city{Moscow}
  \country{Russia}
}

\affiliation{%
  \institution{St. Petersburg Department of the Steklov Institute of Mathematics}
  \city{St. Petersburg}
  \country{Russia}
}

\author{Irina Piontkovskaya}
\email{piontkovskaya.irina@huawei.com}
\orcid{0009-0003-0299-5849}

\affiliation{%
  \institution{Huawei Noah's Ark Lab}
  \city{Moscow}
  \country{Russia}
}

\renewcommand{\shortauthors}{Konstantin Yakovlev et al.}

\newcommand{\todo}[1]{[\textbf{TODO}: #1]}

\begin{abstract}
A recent trend in multimodal retrieval is related to postprocessing test set results via the dual-softmax loss (DSL). While this approach can bring significant improvements, it usually presumes that an entire matrix of test samples is available as DSL input. This work introduces a new postprocessing approach based on Sinkhorn transformations that outperforms DSL. Further, we propose a new postprocessing setting that does not require access to multiple test queries. We show that our approach can significantly improve the results of state of the art models such as CLIP4Clip, BLIP, X-CLIP, and DRL, thus achieving a new state-of-the-art on several standard text-video retrieval datasets both with access to the entire test set and in the single-query setting.
\end{abstract}




\maketitle

\section{Introduction}

Text-video and video-text retrieval have been among key tasks in information retrieval for a long time~\cite{5729374}, with obvious applications to, e.g., video search and indexing. Over the past few years, this field, similar to many in deep learning, has come to be dominated by Transformer-based architectures for producing latent representations, either inherently multimodal such as CLIP~\cite{pmlr-v139-radford21a} or BLIP~\cite{li2022blip} or designed for independent processing of text (e.g. the original Transformer~\cite{NIPS2017_3f5ee243}) and video (usually in the form of individual frames processed by architectures such as Vision Transformer (ViT)~\cite{DBLP:journals/corr/abs-2010-11929}). This progress in backbone architectures has been complemented by new retrieval-specific ideas that explore various cross-modal and differently grained interactions between text and video frames (see Section~\ref{sec:related} for an overview).
One interesting recent trend deals with \emph{postprocessing} the retriever's results.
The most common transformation here is the dual-softmax loss (DSL)~\cite{DBLP:journals/corr/abs-2109-04290} that (when applied as postprocessing) takes as input a matrix of query-result scores and applies a softmax-based transformation whose primary goal is to reduce the scores of ``hubs'', i.e., videos that may appear to be a good answer to a wide variety of queries. However, this kind of postprocessing is ``unfair'' in the sense that a full application requires access to the entire test set, and it is unclear how much of the improvement comes from ``peeking'' into the test set distribution that would be unavailable in a real application.
Our first contribution is a new postprocessing approach based on the \emph{Sinkhorn transformation}, introduced for optimal transport~\cite{NIPS2013_af21d0c9} and then in deep learning~\cite{Mena2018LearningLP}, but not in the IR setting. We find that it consistently outperforms DSL across a variety of video retrieval datasets and models. Second, we propose a new evaluation setup, the \emph{single-query setting}, where postprocessing has access to only one test query, sampling the rest from the training set. We show that while the positive effect of postprocessing is significantly reduced in this setting, some of it remains, and it can thus be achieved in IR practice. The Sinkhorn transformation again fares better in this setting. We also study the \emph{zero-shot single-query setting}, where models are not fine-tuned on the training sets of evaluation corpora, with similar results.
In what follows, Section~\ref{sec:related} surveys related work, Section~\ref{sec:methods} introduces DSL and Sinkhorn postprocessing, Section~\ref{sec:eval} shows our evaluation study,
and Section~\ref{sec:concl} concludes the paper.

\section{Related work}\label{sec:related}

\textbf{Datasets}. 
Several video retrieval datasets have become industrial and academic standards; many works report results for specific train/test splits, enabling easy comparison for retrieval models. Apart from MSR-VTT~\cite{xu2016msr}, MSVD~\cite{chen2011collecting}, DiDeMo~\cite{anne2017localizing}, and ActivityNet~\cite{krishna2017dense} described in Section~\ref{sec:eval}, we note:
\begin{inparaenum}[(i)]
\item LSMDC (Large Scale Movie Description Challenge)~\cite{Rohrbach_2015_CVPR}, designed to provide linguistic descriptions of movies for visually impaired people, with $>$118K short video clips from 202 movies;
\item YouCook2~\cite{DBLP:journals/corr/ZhouXC17}, one of the largest task-oriented datasets of instructional videos with 2000 long videos from 89 cooking recipes, averaging 22 videos per recipe;
\item VATEX~\cite{DBLP:journals/corr/abs-1904-03493}, a multilingual dataset with over 41250 videos and 825000 captions in English and Chinese.
\end{inparaenum}

\begin{figure*}[t]
\includegraphics[width=.75\linewidth]{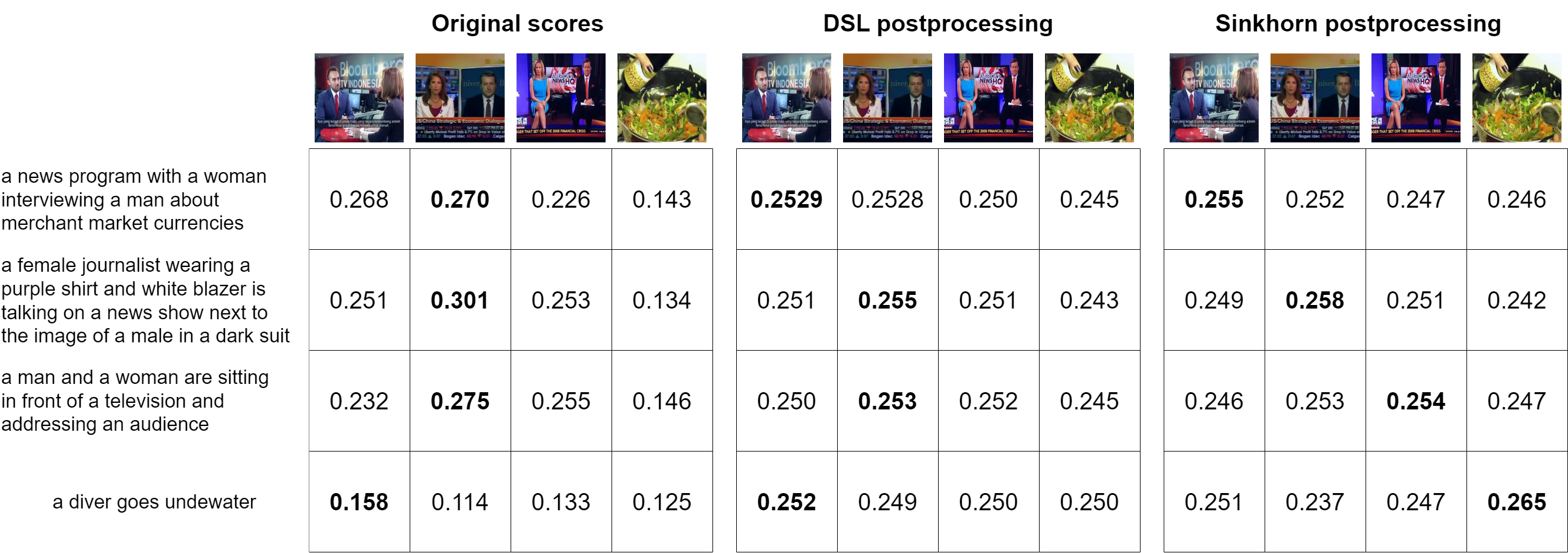}

\caption{Sample results of the DSL and Sinkhorn postprocessing methods for text-video retrieval.}\label{fig:example}
\end{figure*}

\textbf{Models}. Text-video and video-text retrieval models often rely on multimodal embeddings from Transformer-based models such as CLIP~\cite{pmlr-v139-radford21a}, which trains joint text-image representations via contrastive losses and has been extended to videos in \emph{CLIP4Clip}~\cite{luo2022clip4clip},  BLIP~\cite{li2022blip}, and X-CLIP~\cite{https://doi.org/10.48550/arxiv.2207.07285} (Section~\ref{sec:eval}). Among other models we note:
\begin{inparaenum}[(i)]
\item \emph{ClipBERT}~\cite{DBLP:journals/corr/abs-2102-06183}, a Transformer-based framework with end-to-end learning;
\item \emph{VideoClip}~\cite{xu-etal-2021-videoclip} that pretrains a single unified model for zero-shot video and text understanding;
\item \emph{X-Pool}~\cite{gorti2022xpool}, a cross-modal attention approach where a text query attends to semantically similar frames;
\item \emph{Tencent Text-Video Retrieval}~\cite{https://doi.org/10.48550/arxiv.2204.03382} that explores cross-modal interactions organized in several hierarchical layers: token-wise interactions on the word-frame and clip-phrase level and the global dot product on the level of global representations;
\item \emph{LiteVL}~\cite{DBLP:journals/corr/abs-2210-11929} that applies BLIP to video retrieval with temporal attention modules (cf. TimeSformer~\cite{DBLP:journals/corr/abs-2102-05095}) and text-dependent pooling 
that allow to adapt to video-language downstream tasks without pretraining, starting from image-text BLIP.
\end{inparaenum}

\section{Methods}\label{sec:methods}


\textbf{Dual-Softmax Loss} (\textbf{DSL})~\cite{DBLP:journals/corr/abs-2109-04290}. 
%
For a text-to-video similarity matrix $\AA$ and a constant $T > 0$, the dual softmax function is defined as
\[
    \begin{array}{rcl}
        \AA' &=& \mathrm{Softmax}(T\cdot\AA,\, \dim=0),\\
        \mathrm{DualSoftmax}(\AA, T) &=& \mathrm{Softmax}(\AA \odot \AA',\, \dim=1),
    \end{array}
\]

\noindent
where $\odot$ denotes elementwise multiplication. $\mathrm{DualSoftmax}$ decreases the scores for source objects that are relevant to many queries, which leads to more balanced retrieval results.

\begin{table*}[t]
  \caption{Retrieval results with DSL and Sinkhorn postprocessing with full access to the test set, Recall@k metrics.}\label{tbl:res}
  \label{tab:freq}\small
  \begin{tabular}{llccc|ccc|ccc|ccc}
    \toprule
    & & \multicolumn{3}{c}{\textbf{MSR-VTT}} & \multicolumn{3}{c}{\textbf{MSVD}} & \multicolumn{3}{c}{\textbf{DiDeMo}} & \multicolumn{3}{c}{\textbf{ActivityNet}} \\
    & \textbf{Model + Postprocessing} & \textbf{R@1} & \textbf{R@5} & \textbf{R@10} & \textbf{R@1} & \textbf{R@5} & \textbf{R@10} & \textbf{R@1} & \textbf{R@5} & \textbf{R@10} & \textbf{R@1} & \textbf{R@5} & \textbf{R@10} \\
    \midrule
    \multirow{9}{*}{\rotatebox{90}{\textbf{Text-video retrieval}}} &
    CLIP4Clip & 41.9 & 68.6 & 79.8 & 46.2 & 76.1 & 84.9 & 41.5 & 68.8 & 79.1 & 39.1 & 70.8 & 83.1\\
    & $\quad$+DSL & 47.0& 72.6 & 82.2 & 48.5 & 77.9 & 85.9 & 47.4 & 72.1 & 81.7 & 49.4 & 77.5 & 87.2\\
    & $\quad$+Sinkhorn & \textbf{47.9} & \textbf{74.6} & \textbf{83.1} & \textbf{49.0} & \textbf{78.3} & \textbf{86.2} & \textbf{50.2} & \textbf{73.7} & \textbf{82.5} & \textbf{52.0} & \textbf{79.8} & \textbf{88.9} \\
    \cmidrule{2-14}
    & BLIP & 43.3& 65.6& 74.7 & 45.5 & 73.6 & 81.7 & 43.2 & 69.3 & 75.9 & 34.3 & 60.2 & 72.2 \\
    & $\quad$+DSL & 46.3 & 69.1 & 78.2 & 46.9 & 75.2 & 83.3 & 48.0 & 72.6 & 79.4 & 41.3 & 67.2 & 78.2\\
    & $\quad$+Sinkhorn & \textbf{50.2} & \textbf{72.3} & \textbf{81.0} & \textbf{47.7} & \textbf{76.1} & \textbf{84.0} & \textbf{51.0} & \textbf{76.5} & \textbf{82.6} & \textbf{47.2} & \textbf{71.5} & \textbf{81.3} \\
    \cmidrule{2-14}
    & X-CLIP & 46.3 & 73.1 & 81.9 & 47.1 & 77.1 & 85.8 & 44.8 & 73.6 & 82.8 & 42.9 & 73.2 & 84.7 \\
    & $\quad$+DSL & 52.4 & 76.4 & 84.2 & 49.6 & 79.3 & 87.0 & 52.0 & 76.0 & 83.3 & 52.4 & 80.2 & 89.0 \\
    & $\quad$+Sinkhorn & \textbf{53.4} & \textbf{76.5} &\textbf{85.1} & \textbf{50.2} & \textbf{79.8} & \textbf{87.4} & \textbf{55.2} & \textbf{79.0} & \textbf{85.5} & \textbf{55.5} & \textbf{82.0} & \textbf{90.0} \\
    \midrule
    \multirow{9}{*}{\rotatebox{90}{\textbf{Video-text retrieval}}} &
    Clip4Clip & 41.1 & 70.9 & 79.7 & 55.7 & 82.0 & 89.7 & 41.8 & 69.3 & 79.5 & 40.9 & 72.0 & 84.1 \\
    & $\quad$+DSL & 46.9 & 72.8 & 82.5 & 74.0 & 91.9 & \textbf{96.9} & 48.9 & 73.5 & 82.5 & 49.4 & 78.0 & 87.6 \\
    & $\quad$+Sinkhorn & \textbf{48.1} & \textbf{73.9} & \textbf{83.6} & \textbf{76.0} & \textbf{92.5} & 96.5 & \textbf{50.2} & \textbf{73.7} & \textbf{82.5} & \textbf{53.0} & \textbf{79.9} & \textbf{89.3} \\
    \cmidrule{2-14}
    & BLIP & 35.8 & 59.7 & 71.3 & 59.7 & 84.1 & 88.8 & 37.5 & 61.8 & 71.0 & 29.3 & 54.7 & 67.3 \\
    & $\quad$+DSL & 46.5 & 70.5 &  77.9 & \textbf{64.7} & \textbf{86.0} & 90.0 & 49.5 & 72.6 & 80.1 & 42.4 & 66.9 & 76.7 \\
    & $\quad$+Sinkhorn &  \textbf{49.6} & \textbf{71.0} & \textbf{80.3} & 64.2 & \textbf{86.0} &  \textbf{90.6} & \textbf{50.8} & \textbf{74.8} & \textbf{81.4} & \textbf{46} & \textbf{69.9} & \textbf{79.5} \\
    \cmidrule{2-14}
    & X-CLIP & 45.7 & 73.0 & 81.6 & 59.5 & 83.6 & 90.0 & 45.2 & 72.9 & 81.7 & 42.3 & 73.5 & 85.7  \\
    & $\quad$+DSL & 52.2 & 76.4 & 83.9 & 74.3 & 92.4 & 96.4 & 52.6 & 76.6 & 83.6 & 52.4 & 80.1 & 89.2 \\
    & $\quad$+Sinkhorn & \textbf{53.0} & \textbf{77.4} & \textbf{84.6} & \textbf{75.8} & \textbf{93.4} & \textbf{96.8} & \textbf{56.0} & \textbf{78.2} & \textbf{84.8} & \textbf{56.0} & \textbf{82.0} & \textbf{90.4} \\
  \bottomrule
\end{tabular}
\end{table*}

\textbf{Sinkhorn postprocessing}. We propose to use a different idea, the \emph{Sinkhorn layer}, for the same purpose of rescoring retrieval results. It was introduced by \citet{Mena2018LearningLP} as an extension of the Gumbel-Softmax trick~\cite{DBLP:conf/iclr/JangGP17} and later used for grammatical error correction by \citet{Mallinson2022EdiT5ST}; see also \cite{NIPS2013_af21d0c9}. To the best of our knowledge, Sinkhorn layers have not been employed in IR.

For an arbitrary matrix $\mathbf{A}$, a Sinkhorn step is defined as follows:
$$
\begin{array}{rcl}
    \AA' &=& \mathbf{A}^{(k)} - \mathrm{LogSumExp}(\AA^{(k)}, \dim=0),\\
    \AA^{(k+1)} &=& \mathbf{A}' - \mathrm{LogSumExp}(\AA', \dim=1),
\end{array}
$$
where $\AA^{(0)}=\frac{1}{T}\AA$ for some temperature hyperparameter $T>0$ and $\AA^{(k)}$ is the output of the $k$th Sinkhorn step.
The theoretical motivation here is that when the number of steps $k$ tends to infinity, $\mathrm{exp}(\mathbf{A}^{(k)})$ tends to a doubly 
stochastic matrix, i.e., after applying $\arg\,\max$ to every row we obtain a valid permutation that does not point to the same video twice. Similarly to the dual softmax transformation, this means that after a sufficiently large number of Sinkhorn steps, if the video-to-text score is high then the text-to-video score will also be high.
Fig.~\ref{fig:example} shows sample results for DSL and Sinkhorn postprocessing on a real example. Both transformations try to downgrade the second video that appears best for three queries out of four. In this example, DSL succeeds in distinguishing it from the first video (which is the correct answer to the first query) but still leaves it the best for the third query while Sinkhorn postprocessing gets all three correct; the fourth query in Fig.~\ref{fig:example} does not have a correct answer among these videos so its scores are irrelevant, but it still does not hurt postprocessing results.

\textbf{Single-query setting}. One important drawback of IR rescoring methods is that they need access to an entire test set of queries. 
This runs contrary to the usual IR setting where only a single new query is available. Despite this fact, DSL and QB-Norm~\cite{DBLP:conf/cvpr/BogolinCJLA22} have become standard postprocessing techniques; several recent works use them to improve the final results, including, e.g., CLIP-ViP~\cite{https://doi.org/10.48550/arxiv.2209.06430}, Tencent Text-Video Retrieval~\cite{https://doi.org/10.48550/arxiv.2204.03382}, DRL~\cite{https://doi.org/10.48550/arxiv.2203.07111}, and InternVideo~\cite{https://doi.org/10.48550/arxiv.2212.03191}, sometimes improving the state of the art only with postprocessing.
%
To bring postprocessing techniques to a production environment where only one query is available at a time, we propose the \emph{single-query setting} where we still apply postprocessing to an $m\times m$ matrix of queries, but it contains only one test query and $m-1$ queries randomly sampled from the \emph{training} set. In this way, we separate the advantages that come from rescoring itself from the bonus gained by having additional access to the test set query distribution. We suggest this approach as a ``fair'' way to evaluate postprocessing techniques: it somewhat increases the computational costs but can be easily implemented in practical IR applications.

\section{Experimental setup}\label{sec:eval}

\input{tables}

\textbf{Datasets}.
We evaluate on several standard text-video retrieval datasets.
\textbf{MSR-VTT} \cite{xu2016msr} (MSRVideo to Text) is a video understanding dataset with 118 videos for each of 257 queries from a video search engine, 
10K web video clips with 41.2 hours (each clip lasts for about 10 to 32 seconds) and 200K clip-sentence pairs in total, manually annotated via crowdsourcing with $\approx$20 sentences each; each sentence summarizes the entire clip. There are two widely used ways to split MSR-VTT: MSRVTT-7k (7000 videos for training) \cite{miech2019howto100m} and MSRVTT-9k (9000 videos) \cite{gabeur2020multi}; we used the latter.
%
\textbf{MSVD} \cite{chen2011collecting} (Microsoft Research Video Description Corpus) is a classical dataset of one-sentence descriptions for short videos 
(under 10 seconds).
It has $\approx$120K sentences in 35 languages that are parallel descriptions of 2089 video snippets; we used only the English part with 85K descriptions.
\textbf{DiDeMo} \cite{anne2017localizing} (Distinct Describable Moments) consists of 10K Flickr videos annotated with 40K text captions. It was collected for localizing moments in video with natural language queries, and 
it is usually considered for ``paragraph-to-video'' retrieval, concatenating all descriptions, 
since different queries describe different localized moments in a clip.
%
\textbf{ActivityNet} \cite{krishna2017dense} contains 20K \emph{YouTube} videos annotated with 100K sentences, with 10K videos in the training set. It was intended for dense captioning, which involves both detecting and describing (possibly multiple) events in a video; 
we used the common \emph{val1} set with 4.9K videos.

\textbf{Models}.
We test postprocessing on the results of four state of the art models.
\textbf{CLIP4Clip} \cite{luo2022clip4clip} uses CLIP (Contrastive Language-Image Pretraining)~\cite{radford2021learning} that jointly trains image and text encoders to learn latent representations of image-text pairs. CLIP4Clip transfers the knowledge of pretrained CLIP to text-video retrieval with several modifications, including similarity calculation mechanisms and pretraining on a large-scale video-language dataset.
\textbf{BLIP} (Bootstrapping Language-Image Pre-training)~\cite{li2022blip} 
uses three Transformer encoders---one for images and two for text captions---and a Transformer decoder. Another novelty of BLIP is a special training process based on a system for filtering mined vs synthetically generated captions and combines different parts of the dataset on different training stages.
\textbf{X-CLIP}~\cite{https://doi.org/10.48550/arxiv.2207.07285} 
introduced the idea of \emph{cross-grained} contrastive learning, computing the correlations between coarse and fine-grained features in their multi-grained architecture. In text-video retrieval, it means a combination of video-sentence, video-word, frame-sentence, and frame-word interactions, with video framed encoded by ViT~\cite{DBLP:journals/corr/abs-2010-11929}.
\textbf{DRL} (Disentangled Representation Learning)~\cite{DRLTVR2022} introduces a fine-grained cross-modal interaction mechanism based on token-wise interactions and channel decorrelation regularization that minimizes the redundancy in learned representation vectors.

\subsection{Results}

\textbf{Comparing postprocessing approaches}. Table~\ref{tbl:res} shows the main results of our evaluation
for models and datasets from Section~\ref{sec:eval}.
The proposed Sinkhorn postprocessing consistently and often very significantly (cf. p-values) outperforms DSL across all models and datasets. Thus, we recommend to adapt Sinkhorn transformations as a standard postprocessing technique instead of DSL.
Further, Table~\ref{tbl:res} shows the results of video-text retrieval, i.e., searching for text descriptions by video input. All models allow for this modification, and the results again show that in this direction Sinkhorn postprocessing is consistently better than DSL, and both help a lot compared to the original models.

\textbf{Single-query setting}. Table~\ref{tbl:single} shows the results in the single-query setting, where postprocessing cannot be done on test set queries. To alleviate this, we sample additional queries from the training set, calculate their similarity with test set videos, and then sample a smaller number of queries to form a \textit{pseudo-test set} for DSL and Sinkhorn postprocessing. For a fair comparison, both are given the same pseudo-test sets; we show metrics averaged over three random resamples of the pseudo-test set.
%
Results in the single-query setting are consistently worse than with the full test set
(Table~\ref{tbl:res}), validating our hypothesis that much of the postprocessing effect comes from alleviating the domain shift between training and test sets. However, Table~\ref{tbl:single} shows that even in the single-query setting rescoring approaches do help, and in most experiments Sinkhorn postprocessing fares better than DSL.
Table~\ref{tbl:single} also considers the zero-shot single-query setting, where models are not fine-tuned on the training sets of the corresponding datasets (BLIP is shown only here since it is zero-shot by default). The results are predictably much worse, but both the significant positive effect of postprocessing and the advantage of Sinkhorn over DSL persist in this setting.

Results for the DRL base model, both with full access to the test set and in the single-query settings, are shown separately in Table~\ref{tbl:drl} since we have these results only for the MSR-VTT and \emph{DiDeMo}, and the other datasets have not been covered by its original code~\cite{DRLTVR2022}. Table~\ref{tbl:drl} leads to the same conclusions as Tables~\ref{tbl:res} and~\ref{tbl:single}: postprocessing helps in all settings, and the advantage of Sinkhorn over DSL is even more pronounced in the single-query settings for the DRL model than for other models (Table~\ref{tbl:single}).

\section{Conclusion}\label{sec:concl}

In this work, we have presented a new approach to rescoring information retrieval results based on the Sinkhorn transformation, showing that it outperforms the currently popular DSL postprocessing. We have also shown that most of the positive effects of postprocessing can be attributed to having access to the entire test set, but introduce a ``fair'' evaluation setting with access to only a single query and show that even in this setting the effects of postprocessing are significant, and the Sinkhorn approach still outperforms DSL. We suggest the Sinkhorn transformation as a new standard reranking approach and posit that such postprocessing can be applied in IR practice with a positive effect.

\begin{acks}
\end{acks}
The work of Sergey Nikolenko was supported by a grant for research centers in the field of artificial intelligence, provided by the Analytical Center for the Government of the Russian Federation in accordance with the subsidy agreement (agreement identifier 000000D730321P5Q0002) and the agreement with the Ivannikov Institute for System Programming of the Russian Academy of Sciences dated November 2, 2021, No. 70-2021-00142.

\bibliographystyle{ACM-Reference-Format}
\balance
\bibliography{sample-base}





\end{document}

%% file: tables.tex
\begin{table}[t]\centering\setlength{\tabcolsep}{2.25pt}\small
  \caption{Text-video retrieval with DSL and Sinkhorn postprocessing in the single-query setting, Recall@k metrics. Best result in bold; with$^*$~-- p-value $<0.05$, with$^\ddag$~-- p-value $<0.001$.}\label{tbl:single}
  \begin{tabular}{lccc|ccc|ccc}
    \toprule
    & \multicolumn{3}{c}{\textbf{MSR-VTT}} & \multicolumn{3}{c}{\textbf{DiDeMo}} & \multicolumn{3}{c}{\textbf{ActivityNet}} \\
    \textbf{Model} & \textbf{@1} & \textbf{@5} & \textbf{@10} & \textbf{@1} & \textbf{@5} & \textbf{@10} & \textbf{@1} & \textbf{@5} & \textbf{@10} \\
    \midrule
    \multicolumn{10}{c}{\textbf{Text-video retrieval, single-query setting}}\\\midrule
    CLIP4Clip & 41.9 & 68.6 & 79.8 & 41.5 & 68.8 & 79.1 & 39.1 & \textbf{70.8} & \textbf{83.1} \\
    $\ $+DSL & 42.1 & 68.5 & 79.8 & \textbf{42.0}$^*$ & 68.7 & \textbf{79.6}$^*$ & \textbf{39.6}$^*$ & 70.4 & 82.0\\
    $\ $+Sinkhorn & \textbf{42.3} & \textbf{68.9}$^*$ & \textbf{80.4}$^\ddag$ & 41.2 & \textbf{68.9} & 79.1 & 38.5 & 69.7 & 82.2\\
    \midrule
    X-CLIP & 46.3 & \textbf{73.1} & \textbf{81.9} & 44.8 & 73.6 & \textbf{82.8} & 42.9 & 73.2 & 84.7  \\
    $\ $+DSL & 46.0  & 72.8 & 81.3 & \textbf{45.4} & 74.4 & 81.9 & \textbf{43.6}$^*$ & 73.1 & 84.4 \\
    $\ $+Sinkhorn & \textbf{46.4}$^*$ & 72.7 & 81.5 & \textbf{45.4} & \textbf{74.6} & 82.1 & 43.0 & \textbf{73.4} & \textbf{84.8} \\
    \midrule
    \multicolumn{10}{c}{\textbf{Text-video retrieval, zero-shot single-query setting}}\\\midrule
    CLIP4Clip & 30.9 & 54.2 & 63.3 & 29.5 & 53.6 & 64.7 & 21.7 & 46.1 & 60.0 \\
    $\ $+DSL & 34.5 & 57.0 & 67.7 & 31.0 & 57.2 & 69.0 & 24.8 & 50.7 & 64.6 \\
    $\ $+Sinkhorn & \textbf{34.9}$^*$ & \textbf{57.7}$^*$ & \textbf{69.0}$^\ddag$ & \textbf{32.2}$^\ddag$ & \textbf{58.0}$^*$ & \textbf{69.3} & \textbf{26.1}$^*$ & \textbf{53.5}$^\ddag$ & \textbf{67.0}$^\ddag$\\
    \midrule
    BLIP & 43.3 & 65.6& 74.7 & 43.2 & 69.3 & 75.9 & 34.3 & 60.2 & 72.2  \\
    $\ $+DSL & 43.6 & 65.8 & 75.2 & \textbf{43.7} & 70.2 & 78.1 & 36.6 & 63.2 & 74.8 \\
    $\ $+Sinkhorn & \textbf{43.9} &  \textbf{66.9}$^\ddag$ & \textbf{76.3}$^\ddag$ & \textbf{43.7} & \textbf{70.7}$^*$ & \textbf{78.7}$^\ddag$ & \textbf{38.7} & \textbf{65.4} & \textbf{76.6}\\
    \midrule 
    X-CLIP & 31.4 & 53.4 & 63.6 & 29.5 & 54.4 & 63.4 &  21.5 & 45.6 & 59.2 \\
    $\ $+DSL & 34.3 & 57.1 & 68.2 & 31.2 & 58.7 & 69.5 & 25.0 & 50.8 & 64.8 \\
    $\ $+Sinkhorn & \textbf{34.7}$^*$ & \textbf{58.2}$^\ddag$ & \textbf{69.2}$^\ddag$ & \textbf{32.0}$^\ddag$ & \textbf{59.7}$^\ddag$ & \textbf{70.7}$^\ddag$ & \textbf{26.5}$^*$ & \textbf{52.6}$^\ddag$ & \textbf{66.9}$^\ddag$ \\
  \bottomrule
\end{tabular}\vspace{.2cm}

\setlength{\tabcolsep}{5pt}
  \caption{Experimental results with DSL and Sinkhorn postprocessing for the DRL base model, Recall@k metrics. Best result in bold; with$^*$~-- p-value $<0.05$, with$^\ddag$~-- p-value $<0.001$.}\label{tbl:drl} 
  \begin{tabular}{lccc|ccc}
    \toprule
    & \multicolumn{3}{c}{\textbf{MSR-VTT}} & \multicolumn{3}{c}{\textbf{DiDeMo}} \\
    \textbf{Model} & \textbf{@1} & \textbf{@5} & \textbf{@10} & \textbf{@1} & \textbf{@5} & \textbf{@10} \\\midrule
    \multicolumn{7}{c}{\textbf{Text-video retrieval, full access to the test set}}\\
    \midrule
    DRL & 46.5 & 73.1 & 82.4 & 46.6 & 75.9 & 83.1 \\
    $\ $+DSL & 50.5 & 77.6 & 86.5 & 53.1 & 78.0 & 85.1 \\
    $\ $+Sinkhorn & \textbf{52.6} & \textbf{79.5} & \textbf{87.0} & \textbf{55.9} & \textbf{80.5} & \textbf{87.4}\\
    \midrule
     \multicolumn{7}{c}{\textbf{Video-text retrieval, full access to the test set}}\\
    \midrule
    DRL & 43.8 & 74.0 & 83.6 & 45.8 & 74.7 & 83.6 \\
    $\ $+DSL & 50.4 & 77.7 & 86.2 & 54.4 & 79.0 & 86.5 \\
    $\ $+Sinkhorn & \textbf{51.8} & \textbf{77.9} & \textbf{86.3} & \textbf{56.3} & \textbf{79.8} & \textbf{86.8}\\
    \midrule
    \multicolumn{7}{c}{\textbf{Text-video retrieval, single-query setting}}\\\midrule
    DRL & 46.5 & 73.1 & 82.4 & 46.6 & 75.9 & 83.1 \\
    $\ $+DSL & \textbf{47.0}$^\ddag$ & 73.1 & 82.7 & 47.5 & \textbf{76.5}$^*$ & 84.1\\
    $\ $+Sinkhorn & 46.6 & \textbf{73.4}$^\ddag$ & \textbf{83.1}$^\ddag$ & \textbf{47.7} & 76.0 & \textbf{84.4}\\
    \midrule
    \multicolumn{7}{c}{\textbf{Text-video retrieval, zero-shot single-query setting}}\\\midrule
    DRL & 24.3 & 46.8 & 56.1 & 23.9 & 45.2 & 56.0\\
    $\ $+DSL & 26.3 & 51.2 & 60.8 & 26.3 & 52.0 & 65.8 \\
    $\ $+Sinkhorn & \textbf{28.4}$^\ddag$ & \textbf{53.1}$^\ddag$ & \textbf{64.7}$^\ddag$ & \textbf{28.8}$^\ddag$ & \textbf{54.2}$^\ddag$ & \textbf{66.4}$^*$\\
  \bottomrule
\end{tabular}
\end{table}